\pdfoutput=1

\documentclass[11pt]{article}

\usepackage{ACL2023}

\usepackage{times}
\usepackage{latexsym}
\usepackage[T1]{fontenc}
\usepackage[utf8]{inputenc}
\usepackage{microtype}
\usepackage{inconsolata}
\usepackage{enumitem,kantlipsum}
\usepackage{graphicx}
\usepackage{caption}
\usepackage{amsmath}
\usepackage{amssymb}
\usepackage{pgfplots}
\pgfplotsset{compat=newest}
\usepackage{comment}
\usepackage{booktabs}
\usepackage{hyperref}
\usepackage[overload]{empheq}
\usepackage{makecell}
\usepackage{xcolor,colortbl}
\usepackage[ruled,vlined]{algorithm2e}
\usepackage{array,multirow}
\usepackage{comment}
\usepackage{pifont}
\usepackage{url}
\usepackage{siunitx}
\usepackage{xcolor}
\usepackage{soul}
\usepackage{arydshln}
\setlist[itemize]{leftmargin=*}

\newcommand{\todo}[1]{{\textit{\color{magenta}{#1}}}}
\newcommand{\maxim}[1]{{\textit{\color{teal}{#1}}}}

\newcommand{\specialcell}[2][c]{%
  \begin{tabular}[#1]{@{}c@{}}#2\end{tabular}
}

\newcommand{\hlc}[2][yellow]{{%
    \colorlet{foo}{#1}%
    \sethlcolor{foo}\hl{#2}}%
}

\title{Automatic Extraction of Disease Risk Factors from Medical Publications}

\author{
Maxim Rubchinsky$^{1}$\hspace{1.25cm}
Ella Rabinovich$^{1}$\hspace{1.25cm}
Adi Shraibman$^{1}$\hspace{1.25cm}
Netanel Golan$^{3,4}$
\vspace{0.1cm} \\
\textbf{Tali Sahar}$^{4}$\hspace{3cm}
\textbf{Dorit Shweiki}$^{2}$ 
\vspace{0.2cm} \\
$^{1}$\normalsize{School of Computer Science, The Academic College of Tel Aviv-Yaffo, Israel} \\
$^{2}$\normalsize{Bioinformatics, School of Computer Science, The Academic College of Tel Aviv-Yaffo, Israel} \\
$^{3}$\normalsize{Division of Cardiology, Tel-Aviv Sourasky Medical Center, Tel Aviv, Israel} \\
$^{4}$\normalsize{Faculty of Medicine, The Hebrew University of Jerusalem, Israel}
\vspace{0.1cm} \\
\normalsize{\texttt{maxim@rubchinsky.com}, \texttt{\{ellara,adish,dorits\}@mta.ac.il}, 
\texttt{Netanel.golan@mail.huji.ac.il}}
}

\begin{document}
\maketitle
\begin{abstract}
We present a novel approach to automating the identification of risk factors for diseases from medical literature, leveraging pre-trained models in the bio-medical domain, while tuning them for the specific task. Faced with the challenges of the diverse and unstructured nature of medical articles, our study introduces a multi-step system to first identify relevant articles, then classify them based on the presence of risk factor discussions and, finally, extract specific risk factor information for a disease through a question-answering model. 

Our contributions include the development of a comprehensive pipeline for the automated extraction of risk factors and the compilation of several datasets, which can serve as valuable resources for further research in this area. These datasets encompass a wide range of diseases, as well as their associated risk factors, meticulously identified and validated through a fine-grained evaluation scheme. We conducted both automatic and thorough manual evaluation, demonstrating encouraging results. We also highlight the importance of improving models and expanding dataset comprehensiveness to keep pace with the rapidly evolving field of medical research.
\end{abstract}

\section{Introduction}
\label{sec:introduction}

Automatic identification of risk factors for diseases plays a pivotal role in preventive medicine, enabling healthcare professionals to formulate effective prevention strategies and improve patient outcomes. Traditionally, this process has relied heavily on manual review of extensive medical literature, a time-consuming and labor-intensive task, hindering knowledge accessibility and effective usage.

As a concrete example, recently, compelling evidence has emerged linking Lipoprotein A (Lp(a)) --- a particle operating similarly to the more familiar LDL molecule --- to the pathogenesis of atherosclerosis and subsequent coronary artery disease, commonly referred to as Myocardial Infarction (MI). Despite the established role of Lp(a) as a risk factor \citep{kronenberg2022lipoprotein}, many primary care clinicians remain inadequately informed, occasionally lacking knowledge regarding its testing procedures. Moreover, in a conversation with a board-certified professor of interventional cardiology, he disclosed receiving frequent inquiries from other clinicians questioning the necessity of referrals for Lp(a) testing. This highlights the pressing need for an automated tool capable of screening vast amounts of scientific literature and identifying prominent risk factors for various diseases.

Despite significant advances in the field of natural language processing, automatic extraction of disease risk factors from \textit{scientific medical literature} remains a challenging endeavor. Contrary to the analysis of \textit{electronic health records} \citep{chen2015automatic, boytcheva2017identification, chokwijitkul2018identifying}, here the primary challenge lies in the diverse and unstructured nature of medical publications, where risk factors are described in various contexts and formats. What is more, the continuous discovery of new risk factors necessitates a dynamic approach that can adapt to the evolving body of medical knowledge. This study introduces a novel approach to automating the identification of disease risk factors from medical literature. 

\begin{figure*}[h!]
\centering
\resizebox{1.0\textwidth}{!}{
\includegraphics{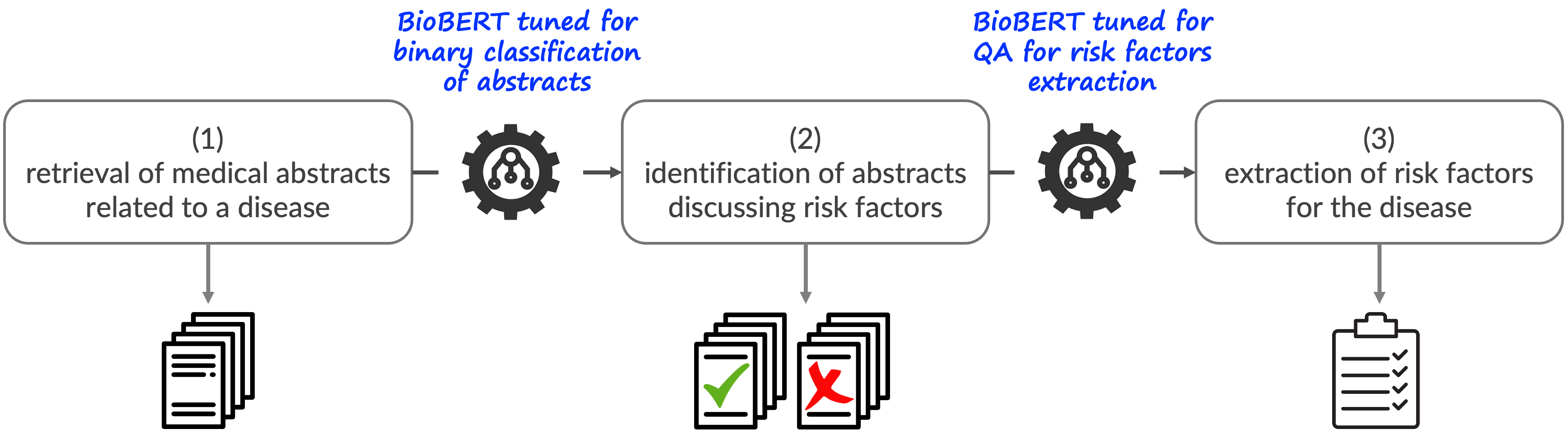}
}
\caption{The pipeline for extraction of disease's risk factors: (1) medical abstracts are retrieved from PubMed, (2) a specifically fine-tuned binary classifier is used to identify abstracts with risk factors information, and (3) precise textual spans containing risk factors are extracted via a QA model, fine-tuned on manually annotated QA items.}
\label{fig:approach}
\end{figure*}

Utilizing pre-trained large language models, based on BioBERT \citep{lee2020biobert}, we developed a multi-step system, that first identifies relevant medical articles, classifies them based on the presence of risk factor discussions, and then extracts specific risk factor information through a question answering (QA) model.
Our approach to extraction of disease risk factors is illustrated in Figure~\ref{fig:approach}: (1) medical abstracts are retrieved from PubMed, (2) a specifically fine-tuned binary classifier is used to identify abstracts with risk factors information, and (3) textual spans containing risk factors are extracted via a question answering model, fine-tuned on manually annotated QA items.

The contribution of this work, therefore, twofold: First, we present a comprehensive pipeline for automated extraction of risk factors. Second, we compile and make available several datasets that can serve as valuable resources for future research in this field. These datasets include a carefully annotated, large and diverse set of over 1,700 risk factors associated with 15 diseases, as well as set of over 160,000 automatically extracted risk factors,\footnote{We note that the set of over 160,000 automatically extracted risk factors are of admittedly mixed quality (see Section~\ref{sec:human-evaluation} and Table~\ref{tbl:final-annotations} for details), yet, we thought this data can serve the community for further research in the field.} with almost 1,500 manually assessed for their quality, using a fine-grained annotation scheme.\footnote{All code and data are available at \url{https://github.com/maximrub/diseases-risk-factors}.} 


We survey the related work in Section~\ref{sec:related-work} and detail on collection and annotation of our datasets in Section~\ref{sec:dataset}. We next describe our approach to the task and report experimental results in Section~\ref{sec:methodology-experiments}. Human evaluation results are presented in Section~\ref{sec:human-evaluation}. Discussion of the difficulty of the task and the limitations of this work are presented in Section~\ref{sec:discussion}. We conclude this study in Section~\ref{sec:conclusions}.

\section{Related Work}
\label{sec:related-work}


Automatic identification of disease risk factors through the analysis of medical texts has garnered interest across various research domains, particularly in applying natural language processing and machine learning techniques to electronic health records (EHRs) and electronic medical records (EMRs).  Here we review key contributions in this area, highlighting approaches that parallel and diverge from our focus on free-text medical articles.

\citet{chokwijitkul2018identifying} explore the utilization of deep learning models to extract heart disease risk factors from EHRs. The approach, grounded in analyzing structured data within EHRs, contrasts with our exploration of unstructured text in medical literature, underscoring the diversity in data sources for risk factor identification. 
\citet{boytcheva2017identification} attempt at mining clinical texts for risk factor identification using association rules. Specifically, they handle data in XML format from the Diabetes Register, indicating a structured approach to data analysis. This work differs from ours in terms of both data source type (clinical narratives), as well as in our broader application to unstructured, free-text medical articles and the use of pre-trained large language models (LLMs) for the task of text understanding.


A comprehensive work on identifying risk factors for heart disease (from clinical data) over time was done in a shared task organized by UTHealth\footnote{The University of Texas Health Science Center.} \citep{stubbs2015identifying}.
\citet{sheikhalishahi2019natural} offer an overview of NLP applications in analyzing clinical notes for chronic disease management, highlighting the increasingly significant contribution of language models to healthcare applications. In the domain of precision medicine, \citet{sabra2017semantic} focus on extracting semantic information and assessing sentiments in clinical notes.

Various works have employed data mining and machine learning (ML) techniques for identifying risk factors from patient data \citep{abdelhamid2023identification}, or clinical outcome prediction \citep{kavakiotis2017machine, mehmood2021prediction, naik2021literature}. Recently, the identification of risk factors for delirium prediction, a rare adverse reaction observed in COVID-19 patients, was developed utilizing ML applied to nursing records \cite{miyazawa2024identification}. 
Additional line of studies focuses on building language models specifically-tailored for medical literature related tasks \citep{roitero2021dilbert, yang2022large, singhal2023towards}.


Several significant contributions have been made in the field of biomedical relation extraction, which includes identifying factors that predispose individuals to diseases. The SemRep \cite{kilicoglu2020broad} tool extracts semantic predications from biomedical texts, including relationships such as "predisposes". The outputs of SemRep have been used to create SemMedDB \cite{kilicoglu2012semmeddb}, a large-scale repository of semantic predications from PubMed. Building on these resources, BioPREP \cite{hong2021bioprep} employs deep learning techniques for predicate classification. The BioRED \cite{luo2022biored} dataset includes a "positive correlation" relation between diseases and other biomedical entities like genes and chemicals.

\paragraph{Conclusion}
While the majority of existing research focuses on analyzing structured \textit{electronic health records} and \textit{electronic medical records} to identify disease risk factors, our study pushes beyond these confines by examining free-text medical literature. Processing unstructured medical text introduces distinct challenges, especially due to language complexity, variation, and the potential for nuanced double meanings, and even worse, due to the necessity to discern context accurately. Consequently, it opens up expansive opportunities for subtle understandings of disease risk factors, facilitating both research and practical applications. 

\section{Dataset}
\label{sec:dataset}


Data collection process for this work can be viewed as a three-step process: (i) collection of the set of disease names spanning multiple disease families, (ii) manual annotation of scientific article abstracts containing explicit mention of risk factors of a subset of diseases -- "abstracts seed", and (iii) manual annotation of risk factors description (span) in abstract texts found in (ii) -- "risk factors seed". We detail on each step in this multi-phase procedure.

\subsection{Disease Dataset Collection}
Aiming to assemble a comprehensive list of diseases, we made use of the KEGG Disease Database API\footnote{KEGG database: \url{https://www.kegg.jp/kegg/disease/}; specifically, we used its REST API service at \url{https://www.kegg.jp/kegg/rest/} for retrieval.} to retrieve disease-related information, including names, description and relevant medical codes such as MeSH (Medical Subject Headings), ICD-10 and ICD-11.\footnote{As of April 2024, ICD-11 (International Classification of Diseases, v11) is the most up-to-date code collection.} This process resulted in 2,624 distinct disease names, comprising the foundation for further retrieval of scientific abstracts and, ultimately, automatic extraction of risk factors, from scientific medical literature.

\subsection{Seed Dataset with Relevant Abstracts}
\label{sec:abstracts-retrieval}

\paragraph{Retrieval of Abstracts Discussing Risks}
Using the list of disease names retrieved from KEGG, we next queried PubMed\footnote{\url{https://pubmed.ncbi.nlm.nih.gov}} --- a large, reliable, and authoritative resource of biomedical literature --- for article abstracts containing the disease names. Specifically, we used the Entrez Programming Utilities\footnote{\url{https://www.ncbi.nlm.nih.gov/books/NBK25501}} via the biogo package.\footnote{\url{https://github.com/biogo/ncbi}} The inherent limitation of this study is related to the fact that only abstracts are freely available through the PubMed interface. However, paper abstracts typically contain a concise summary and main findings of the work, hence constitute a sufficient input for the task at hand. Similarly, prior studies analyzed abstracts retrieved from PubMed for building a biological network \citep{chen2004content}, topical clustering \citep{david2012clustering}, and  identification of negative and positive domain-specific medical terms \citep{vinkers2015use}.

Aiming at retrieval of abstracts discussing findings related to risk factors, we queried PubMed for containment of the phrase "risk factor" in a paper's information: title, abstract or MeSH terms. The following pseudo-code was used for this purpose: 


\begin{figure}[h!]
\centering
\resizebox{1.0\columnwidth}{!}{
\includegraphics{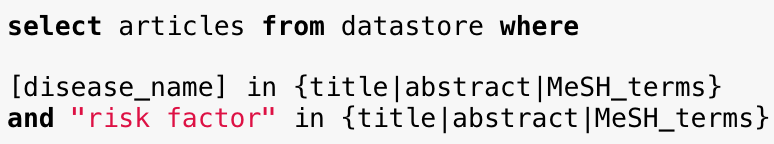}
}
\end{figure}

\noindent where \texttt{disease\_name} refers to the disease we are seeking risk factors for, and the exact search term "risk factor" (surfacting also the plural "risk factor\textbf{s}") can appear in abstract, title or MeSH terms.

\paragraph{Annotation of Abstracts for Risk Factors}
Despite the evident potential, not every abstract with explicit mention of "risk factor" or marked with a "risk factor" MeSH term contains risk factors for a pre-defined disease. As a concrete example, a medical study can mention a list of potential risk factors tested, without any of them showing as significant. We, therefore, define our first (pre-processing) task as automatic classification of a retrieved abstract for spelling out an artifact, found to be a risk factor for the disease in the study.

A qualified annotator with  medical background (one of the authors of this paper) annotated a random set of 182 abstracts. The procedure resulted in 87 positive abstracts (explicitly mentioning a risk factor) and 95 negative, thereby comprising a sufficient training set for the binary classifier -- step (2) in the pipeline in Figure~\ref{fig:approach}. Table~\ref{tbl:abstracts-examples} shows two examples of relevant abstract parts containing risk-related phrases which do or do not qualify as risk factors, as identified by the annotator. Evidently, the nuanced language used to discuss risks in various contexts renders the task as non-trivial for both humans and automatic tools.

\begin{table*}[hbt]
\centering
\resizebox{\textwidth}{!}{
\begin{tabular}{p{17cm}}
article title: \textbf{\textcolor{blue}{Risk Factors} for Pediatric Human Immunodeficiency Virus-related Malignancy} (2003) \\ \hline
\textbf{Context:} Although cancers occur with increased frequency in children with human immunodeficiency virus (HIV) infection, the specific clinical, immunological, and viral \textcolor{blue}{risk factors} for malignancy have not been identified. \textbf{Objective:} To identify \textcolor{blue}{risk factors} for malignancy among HIV-infected children. [...]
\hlc[yellow!50]{Epstein-Barr virus viral load of more than 50 viral genome copies per 105 peripheral blood mononuclear cells was strongly associated with cancer risk} but only for children with CD4 cell counts of at least 200/ microL (odds ratio [OR], 11.33; 95\% confidence interval [CI], 2.09-65.66, P<.001). [...]

\hlc[yellow!50]{High viral burden with EBV was associated with the development of malignancy in HIV-infected children although the effect was modified by CD4 cell count.} The pathogenesis of HIV-related pediatric malignancies remains unclear and other contributing \textcolor{blue}{risk factors} can be elucidated only through further study.

\\ \hline\hline

article title: \textbf{Profound Hypoglycemia and High Anion Gap Metabolic Acidosis in a Pediatric Leukemic Patient Receiving 6-Mercaptopurine} (2024) \\ \hline
A 13-year-old male undergoing maintenance chemotherapy with methotrexate and 6-mercaptopurine (6MP), for very high-risk B-cell acute lymphoblastic leukemia (ALL), presented with vomiting due to severe hypoglycemia with metabolic acidosis. While his laboratory values were concerning for a critically ill child, the patient was relatively well appearing. Hypoglycemia is a rare but serious side effect of 6MP with an unexpectedly variable presentation; therefore, a high index of suspicion is needed for its prompt detection and treatment. [...]

6MP-induced hypoglycemia can be ameliorated with the addition of allopurinol to shunt metabolism in favor of the production of therapeutic metabolites over hepatotoxic metabolites.  Additionally, a morning administration of 6MP and frequent snacks may also help to prevent hypoglycemia. 
Overall, this case adds to the literature of unusual reactions to 6MP including hypoglycemia in an older child without traditional \textcolor{blue}{risk factors}.
\\

\end{tabular}
}
\caption{Article abstracts discussing risk factors (retrieved per the query in Section~\ref{sec:abstracts-retrieval}). Top -- abstract identified as relevant for risk factors extraction by the annotator, where the highlighted part refers to the discussed factor. Bottom -- abstract mentioning "risk factors", yet annotated as irrelevant.}
\label{tbl:abstracts-examples}
\end{table*}

\begin{table*}[hbt]
\centering
\resizebox{\textwidth}{!}{
\begin{tabular}{p{17cm}}
disease: \textbf{Diabetes in Men} \\ \hline
OBJECTIVE: To examine the association between smoking, alcohol consumption, and the incidence of non-insulin dependent diabetes mellitus in men of middle years and older. [...]
%
RESULTS: During 230,769 person years of follow up 509 men were newly diagnosed with diabetes. After controlling for known risk factors \hlc[yellow!50]{men who smoked 25 or more cigarettes daily had a relative risk of diabetes} of 1.94 (95\% confidence interval 1.25 to 3.03) compared with non-smokers. Men who consumed higher amounts of alcohol had a reduced risk of diabetes (P for trend < 0.001). Compared with abstainers men who drank 30.0-49.9 g of alcohol daily had a relative risk of diabetes of 0.61 (95\% confidence interval 0.44 to 0.91). CONCLUSIONS: \hlc[yellow!50]{Cigarette smoking may be an independent, modifiable risk factor for non-insulin dependent diabetes mellitus}. Moderate alcohol consumption among healthy people may be associated with increased insulin sensitivity and a reduced risk of diabetes.
\\ \hline\hline

disease: \textbf{Breast and Colorectal Cancer} \\ \hline
BACKGROUND: Increasing \hlc[yellow!50]{evidence suggests that diabetes mellitus (DM) is associated with increased cancer incidence and mortality}. Several mechanisms involved in diabetes, such as promotion of cell proliferation and decreased apoptosis, may foster carcinogenesis. This study investigated the association between DM and cancer incidence and cancer-specific mortality in patients with breast and colorectal carcinoma. [...]
%
The overall HR for breast cancer incidence was 1.23 (95 per cent confidence interval 1.12 to 1.34) and that for colorectal cancer was 1·26 (1·14 to 1·40) in patients with DM compared with those without diabetes. The overall HR was 1.38 (1.20 to 1.58) for breast cancer- and 1.30 (1.15 to 1.47) for colorectal cancer-specific mortality in patients with DM compared with those without diabetes. CONCLUSION: This meta-analysis indicated that \hlc[yellow!50]{DM is a risk factor for breast and colorectal cancer}, and for cancer-specific mortality.
\end{tabular}
}
\caption{Example of two paper abstracts manually annotated for risk factors. The highlighted text spans (comprising the factors) where marked by the co-author with medical background. Note that in some cases the precise name of the risk factor (e.g., "cigarette smoking") for a disease (e.g., "diabetes in men") is annotated in its broader context, to ensure the model is trained to extract risk factors tied to the disease, and not other, unrelated, artifacts.}
\label{tbl:risk-factors-examples}
\end{table*}

\subsection{QA Seed Dataset with Risk Factors}
\label{sec:rf-annotation}
Given an article abstract specifying a risk factor(s) for a certain disease, we cast the risk factor identification problem as \textit{extractive question answering} scenario, where given the abstract and the question "\texttt{What are the risk factors for \{disease name\}?}", a textual span, containing the answer, will be identified. In Section~\ref{sec:rf-detection} we make use of the established and popular BERT-based QA model -- BioBERT\footnote{\url{https://huggingface.co/dmis-lab/biobert-v1.1}} \citep{lee2020biobert}, and fine-tune it for the task at hand using a manually annotated set of QA items: context (article abstract), a targeted question of the form mentioned above, and a set of manually marked answers in the form \texttt{span\_start} and \texttt{answer\_text} (implying \texttt{span\_end}).

In the absence of suitable annotated datasets for this nuanced task, we developed a web interface for medical students to manually annotate article abstracts. This interface is used for (manual) identification of text segments within abstracts, given the disease discussed in the article. We present a few screenshots of the annotating tool in Appendix~\ref{sec:system-overview}, and release the tool for the community. 

The annotator with medical background marked text spans containing risk factors in a random set of 668 abstracts identified to contain explicit mention of a risk factor,\footnote{The abstracts were sampled from the set automatically classified as "positive" (see Section~\ref{sec:rf-classification})}. resulting in the total of 1,712 QA items, spanning 15 diverse diseases,\footnote{Appendix~\ref{sec:annotated-diseases} reports the full list of diseases.} where each QA item reflects a single risk factor in an abstract that (possibly) encompasses multiple valid risks. Sentences suggesting risk factors significant only within specific population subgroups were denoted as such. Table~\ref{tbl:annotation-examples} presents two examples of QA items: disease name, abstract, and the highlighted risk factor span, as marked by the annotator.

Collectively the carefully-curated and annotated set of abstracts for binary classification of medical articles, and the set of QA items, comprise a high-quality collection for tuning pre-trained language models for the purpose of this study.

\section{Methodology and Experiments}
\label{sec:methodology-experiments}

We further describe in detail our methodological approach, experimental setup and results.

\subsection{Methodology}



As illustrated in Figure~\ref{fig:approach}, we apply a multi-step approach to automate the identification of disease risk factors from medical literature. Central to our methodology is the use of BioBERT, a variant of BERT pre-trained on biomedical texts, enabling nuanced understanding of complex medical language \cite{lee2020biobert}. We next provide details on each step in the process. This model was chosen due to its proven benefits in the biological domain, and its encoder-based architecture -- (arguably) the most appropriate choice for both the classification and extractive question answering tasks at hand.\footnote{Our future work includes investigation of decoder-based models (e.g., GPT), casting the QA part as an abstractive task.}


\subsubsection{Detection of Abstracts with Risk Factors}
\label{sec:rf-classification}
The pre-trained BioBERT-based classifier\footnote{\url{https://huggingface.co/dmis-lab/biobert-v1.1}. We used the default settings with max\_input\_length of 512 tokens, training the classifier for three epochs.} was tuned for abstracts classification using the training part (80\%) out of over 182 manually annotated abstracts (see Section~\ref{sec:abstracts-retrieval}), and tested on the held-out part (20\%), achieving the accuracy of 92\%. Table~\ref{tbl:classification-results} reports the per-class classification results. This encouraging result facilitated our efforts of analyzing content that is most likely to yield valuable insights into disease-risk factor associations. 

\begin{table}[hbt]
\centering
\begin{tabular}{l|c|c|c}
class & P & R & F1 \\ \hline
POS (with risk factor) & 0.89 & 0.94 & 0.92 \\
NEG (w/o risk factor)  & 0.94 & 0.89 & 0.92 \\
\end{tabular}
\caption{Classification results reported on the test set (20\%) of the manually annotated 182 abstracts.}
\label{tbl:classification-results}
\end{table}

We collected a substantial dataset of abstracts, by querying PubMed for each one of over 2400 diseases, as detailed in Section~\ref{sec:abstracts-retrieval}; this step resulted in 137,740 abstracts. We next apply the fine-tuned classifier to identify abstract potentially containing risk factors for a disease. Out of the total number of 137,740 abstracts, 89,834 were classified as positive -- containing explicit mentions of risk factors for diseases. Naturally, some diseases (and disease families) resulted in more prolific retrieval, due to their higher coverage in the medical literature: while various cancer types (e.g., Carcinoma, Leukemia) have large body of related articles, genetic disorders are surveyed less frequently in the context of risk factor discussion.


\subsubsection{Identification of Disease Risk Factors}
\label{sec:rf-detection}
The collection of abstracts classified positively to contain a risk factor, was then subject to the task of risk factor extraction -- step (3) in Figure~\ref{fig:approach}. 
We cast the task as extractive QA, where the medical abstract represents the context, and the question template is formulated as "\texttt{What are the risk factors for \{disease name\}?}". We anticipate the BioBERT QA model \citep{lee2020biobert} to identify span(s) in the abstract containing the answer (or answers, in case multiple risk factors are mentioned in the same abstract), similarly to examples presented in Table~\ref{tbl:risk-factors-examples}. We fine-tune the model for the specific task, as described below.

\paragraph{Fine-tuning the QA Model} We tuned the BioBERT model for our usecase using the training part (80\%) of the 1,712 QA items annotated manually by the author with medical background (see Section~\ref{sec:rf-annotation}); the remaining 20\% were used for testing. Notably, the set of 15 diseases in the 668 abstracts was carefully split into training and test sets, so that the same disease does not appear in both sets, facilitating the assessment of the model's generalizability and performance across a variety of disease contexts.
%
The model tuning was done using the maximum context length of 384 tokens, learning rate of 2e{-}5, and 25 epochs.

We use two common metrics for automatic evaluation of extractive question answering: \texttt{exact-match} and \texttt{F1-score}. Applied on the test set (342 QA items), the metrics obtained 61.76\% for \texttt{exact-match}, and 88.23\% for \texttt{F1-score}, highlighting the potential of the approach.

\paragraph{Determining the Maximum Answer Length}
We determined the maximum length for answers in our QA model by analyzing the lengths of all answers within our training dataset. We calculated the length of each answer (in characters) and studied their distribution. The maximum answer length was set at the 95th percentile of these lengths to encompass the majority of real-world answers while excluding outliers. This threshold is crucial for maintaining focus on concise and relevant answer segments, thereby enhancing the model's training and operational effectiveness. In practice, when the model evaluates potential answers, it only considers text segments whose length does not exceed this predefined limit. Specifically, the text extracted between the predicted start and end indices is compared against the maximum length, and any text exceeding this threshold is disregarded. 

\paragraph{Identification of Risk Factors at Scale}
Utilizing the fine-tuned QA model, we then processed the collected abstracts to identify and validate risk factors for a wide range of diseases, culminating in a dataset that catalogs these findings in much detail. As a concrete example, the entry for the "B-cell acute lymphoblastic leukemia" includes 16 (not necessarily unique) automatically extracted risk factors. Along with the extracted span, the BioBERT QA model provides its probability (confidence, in the 0-1 range) for the identified answer. For a given disease, we only considered answers exceeding the confidence of \texttt{0.6*max\_answer\_probability}, where the \texttt{max\_answer\_probability} is the maximum probability assigned to an answer for the disease. The final dataset encompasses the total of 162,409 identified risk factors spanning 744 diseases, extracted from 54,820 PubMed abstracts.

Due to the inherently strict nature of the \texttt{exact-match} metric, we could observe multiple cases where the extracted answer was largely correct, but didn't represent a precise overlap with the "gold" answer due to a single missing or redundant word. In particular, while some cases surface useful information about a disease risk factors, they are marked as inaccurate by the automatic metric. We complement the evaluation pipeline by sampling a large amount of (automatically identified) risk factors for diseases, and performing fine-grained human assessment of the results' quality.

\section{Human Evaluation}
\label{sec:human-evaluation}
We next manually evaluated a random sample of 1,485 extracted risk factors spanning 29 various diseases (constituting roughly 1\% of the full set of extracted factors), based on their validity and relevance to the disease in question. 

\subsection{Evaluation Scheme}   
We designed a specifically-tailored, four-tiered annotation scheme for the sake of reliable and accurate evaluation, as detailed below. Each risk factor was scored with one of three annotation marks, following the below annotation scheme:

\paragraph{(1) Valid risk factor for the specified disease:} Correctly identified risk factor extracted for the disease of interest, i.e., the disease in the question introduced to the QA system.

\paragraph{(2) Valid risk factor for a different disease:} Correctly identified risk factor for a different disease, i.e., not the disease in the question introduced to the QA system, indicating capabilities yet highlighting challenges in specificity.

\paragraph{(3) Invalid risk factor:} Phrases and terms that are not considered medical risk factors.

\vspace{0.1in}
Additional distinction was done within the first group (valid risk factor), annotating risk factors with strong statistical correlation, as evident from the abstract by inspecting statistical measurements as odd ratio (OR), and confidence intervals (CIs) -- metrics often used in medical literature for testing the significance of findings, such as the presence of a factor in one population but not the other. 41 out of the total of 1,485 were marked as \textit{highly significant} risk factors; we release these annotations as well to facilitate further research in the community.


\subsection{Evaluation Results}
Table~\ref{tbl:annotation-examples} presents error analysis of correctly- and incorrectly-identified risk factor examples (the first two rows), as well as an example for artifact that does not constitute a risk factor (the last row).

We attribute most factors erroneously annotated with type 3 annotation --- not a risk factor --- to cases where the QA model was required to extract a risk factor from an abstracts that does not contain one. Since the model was trained (and fine-tuned) to \textit{always} identify an answer span for a given context and question, it is expected to yield (admittedly) weak performance on a context lacking the factors at the first place. Notably, a relatively small amount of all manually evaluated examples (around 8.5\%) fall into this category.

Table~\ref{tbl:final-annotations} further summarizes the evaluation results by disease family. The prevalence of type 1 and 2 annotations illustrates the model's effectiveness in identifying risk factors, yet also underscores the challenges in achieving precise disease-specific accuracy. The presence of type 3 annotations, although significantly lower, highlights the ongoing need for the classification model refinement to enhance both specificity and accuracy. 

\paragraph{Error Analysis}
Additional observation can be made about error distribution between type 1 and 2 annotations within and across disease families. Evidently, while some disease families show a balanced ratio between type 1 and 2 annotations (e.g., Infection, Leukemias), others resulted in more mis-identified factors -- type 2 annotation (e.g., Metabolic disorders). We hypothesize that abstracts concerning diseases with a significant, sometimes absolute, genetic component are less likely to address other contributing factors. Consequently, research in this area predominantly focuses on stratifying potential risks for other diseases in individuals already affected by the genetic disorder.

\begin{table*}[hbt]
\centering
\resizebox{\textwidth}{!}{
\begin{tabular}{p{1.75cm}|p{15cm}|c}
disease & abstract excerpt (identified risk factor highlighted) & marker \\ \hline
Chronic Myeloid Leukemia & [...]
RESULTS: \hlc[yellow!50]{Previous diagnoses of dyspepsia, gastritis or peptic ulcers, as well as previous proton pump inhibitor (PPI) medication, were all associated with a significantly increased risk of CML} (RRs, 1.5-2.0; P = 0.0005-0.05). Meanwhile, neither inflammatory bowel disease nor intake of NSAIDs were associated with CML, indicating that it is not gastrointestinal ulcer or inflammation per se that influences risk. [...]
 & 1 \\ \hline
 
Cystic Fibrosis & BACKGROUND: \hlc[yellow!50]{Cystic fibrosis}, like other chronic diseases, \hlc[yellow!50]{is a risk factor for the development of elevated symptoms of depression and anxiety}. [...]
\hlc[yellow!50]{Patient anxiety (OR 2.33) and depression} (OR 4.09) were significantly associated with forced expiratory volume in one second (FEV1) <40\% and forced vital capacity (FVC) <80\% (OR 1.60 and 1.61, respectively). CONCLUSIONS: \hlc[yellow!50]{Cystic fibrosis increases the risk of developing anxiety and depression} in female patients and in mothers. 
 & 2 \\ \hline

Renal Cell Carcinoma & 
RESULTS: A total of 888 incident RCCs and 356 RCC deaths were identified. In models including adjustment for body mass index and energy intake, \hlc[yellow!50]{there was no higher risk of incident RCC associated with consumption of juices} (HR per 100 g/day increment = 1.03; 95\% CI, 0.97-1.09), total soft drinks (HR = 1.01; 95\% CI, 0.98-1.05), [...]
CONCLUSIONS: \hlc[yellow!50]{Consumption of juices or soft drinks was not associated with RCC incidence or mortality after adjusting for obesity.}
 & 3 \\
\end{tabular}
}
\caption{Examples for automatic identification of risk factors in medical abstracts, marked by the annotator:\\1 (valid risk factor for the specified disease) -- stomach diseases are risk factors for CML; 2 (valid risk factor for a different disease) -- CF, the disease of interest, was found to be a risk factor for depression and anxiety; and 3 (not a risk factor) -- juices were \textbf{not} identified as a risk factor for RCC.}
\label{tbl:annotation-examples}
\end{table*}

\vspace{0.2in}
\begin{table*}[hbt]
\centering
\resizebox{\textwidth}{!}{
\begin{tabular}{l|S[table-format=3.0]|S[table-format=3.0]|S[table-format=3.0]|S[table-format=3.0]}  
family (sub-family) & \multicolumn{1}{c|}{\specialcell{(1) valid risk factor \\for the specified disease}} & \multicolumn{1}{c|}{\specialcell{(2) valid risk factor \\for a different disease}} & \multicolumn{1}{c|}{(3) not a risk factor} & \multicolumn{1}{c}{total in family} \\ \hline
Carcinomas & 317 & 285 & 60 & 662 \\
Infection & 45 & 51 & 6 & 102 \\
Leukemias & 208 & 192 & 46 & 446 \\
Lymphomas & 27 & 12 & 4 & 43 \\ \hdashline
Metabolic disorders (GD) & 4 & 60 & 8 & 72 \\
Mucus malefunction (GD) & 11 & 34 & 2 & 47 \\
Cardiomyopathy & 5 & 23 & 0 & 28 \\ \hdashline
Sarcomas & 15 & 5 & 1 & 21 \\
other hematological disorders  & 30 & 32 & 2 & 64 \\ \hline
total                   & 662   & 694   & 129   & 1485  \\
\end{tabular}
}
\caption{Distribution of manual evaluation annotations by disease family. "GD" denotes "genetic disorder". Note the much high number of risk factors identified for common (and potentially fatal) diseases, due to the vast body of empirical literature. The numbers refer to the total number of (not necessarily unique) risk factors identified for a disease family. We hypothesize that abstracts concerning diseases with a significant, sometimes absolute, genetic component are less likely to address other contributing factors; between the dashed lines in the table. 
}
\label{tbl:final-annotations}
\end{table*}

\section{Discussion and Limitations}
\label{sec:discussion}


Our study, while contributing valuable insights into the automation of risk factor identification from medical publications, is subject to several limitations that merit a thorough discussion.

One of the primary limitations is the challenge of accurately distinguishing risk factors specifically associated with the disease in question (type 1) from valid risk factors that are not directly related to the disease under investigation (type 2). While our models demonstrated a high capacity for identifying potential risk factors, the precision in contextualizing these factors to specific diseases varied. This aspect highlights a critical area for future research, emphasizing the need for enhanced specificity in the models to improve their utility in targeted medical research and practice.

Moreover, the study's reliance on free-text medical articles introduces variability in the data quality and representation. The unstructured nature of these texts and the diversity in how risk factors are described pose significant challenges for both the binary classification and question-answering models. Efforts to standardize data representation and improve model robustness against such variability are essential steps forward.


The datasets used in this study, while extensive, are not exhaustive. The landscape of medical research is continuously evolving, with new findings emerging regularly. The datasets, therefore, represent a snapshot in time, and ongoing efforts to update and expand these resources are necessary to maintain their relevance and utility.

Finally, the study's scope was constrained by the computational resources available. Future work could explore more complex models or ensemble approaches that might offer improved accuracy but require more substantial computational power.

Despite these limitations, this study represents a significant step toward automating the identification of disease risk factors from medical literature. Acknowledging and addressing these limitations in future research will be crucial for advancing the field and enhancing the practical applicability of these technologies in healthcare.

\section{Conclusions and Future Work}
\label{sec:conclusions}
This study presented an approach to identifying and extracting disease risk factors from free-text medical articles using advanced natural language processing techniques, specifically leveraging the capabilities of the pre-trained BioBERT-based architecture. Our methodology involved a multi-step process, including the retrieval of relevant articles, binary classification to filter articles discussing risk factors, and a question-answering model to extract specific risk factor information.

We have demonstrated the potential of language technologies to significantly enhance the efficiency and effectiveness of risk factor identification in medical literature. Our contributions to this field are twofold: the presentation of an automated pipeline for risk factor extraction and the creation of valuable datasets for future research.
While our study marks an advancement in the automated extraction of risk factors from medical literature, there remain several avenues for future research and development. Our future directions include introducing improvements to QA model's accuracy and specificity, integration of additional data sources, and evaluation of more advanced LLMs for the task of risk factors identification.

Furthermore, inspired by recent findings that automatic annotations generated by models like GPT-4 can achieve results comparable to human annotations, we plan to investigate the use of GPT-4 for the task of risk factors annotation, and compare its performance with human experts.

\section{Ethical Considerations}
\label{sec:ethical}



We make use of publicly available data in the domain of healthcare, that have been broadly used in numerous studies. Manual annotations were conducted by one of the authors of the paper, with medical background. Due to the required expertise and the inherent difficulty of the task, the mean hourly rate for the annotator was much higher than the established minimum wage.

\section*{Acknowledgements}
We are grateful to Noga Shraibman for much help creating datasets during the early stages of this work. We are also thankful to our three anonymous reviewers for their constructive feedback.

\bibliographystyle{acl_natbib}
\bibliography{custom}

\appendix
\section{Appendices}
\label{sec:appendices}


\subsection{Overview of the Risk Factor Annotation System Architecture}
\label{sec:system-overview}
The risk factor annotation system comprises three main components designed to streamline the process of annotating risk factors in medical articles. This system was instrumental in creating the datasets used in our research.

\paragraph{GraphQL Server}
The backbone of the system is a GraphQL server, which serves as the central communication hub. Hosted on Kubernetes (k8s) for scalability and reliability, the server facilitates data exchange between the user interface and the database. It handles requests for data retrieval and submission, ensuring that the web application and the code can access and store data efficiently.

\paragraph{Web UI}
The front end of the system is a React-based web application, also deployed on Kubernetes for high availability. This intuitive user interface allows medical students and researchers to interact with the system, including retrieving medical articles, annotating risk factors within texts, and submitting these annotations back to the server. The design prioritizes ease of use to facilitate accurate and efficient annotation work.

\begin{figure*}
\centering
\includegraphics[width=1\textwidth]{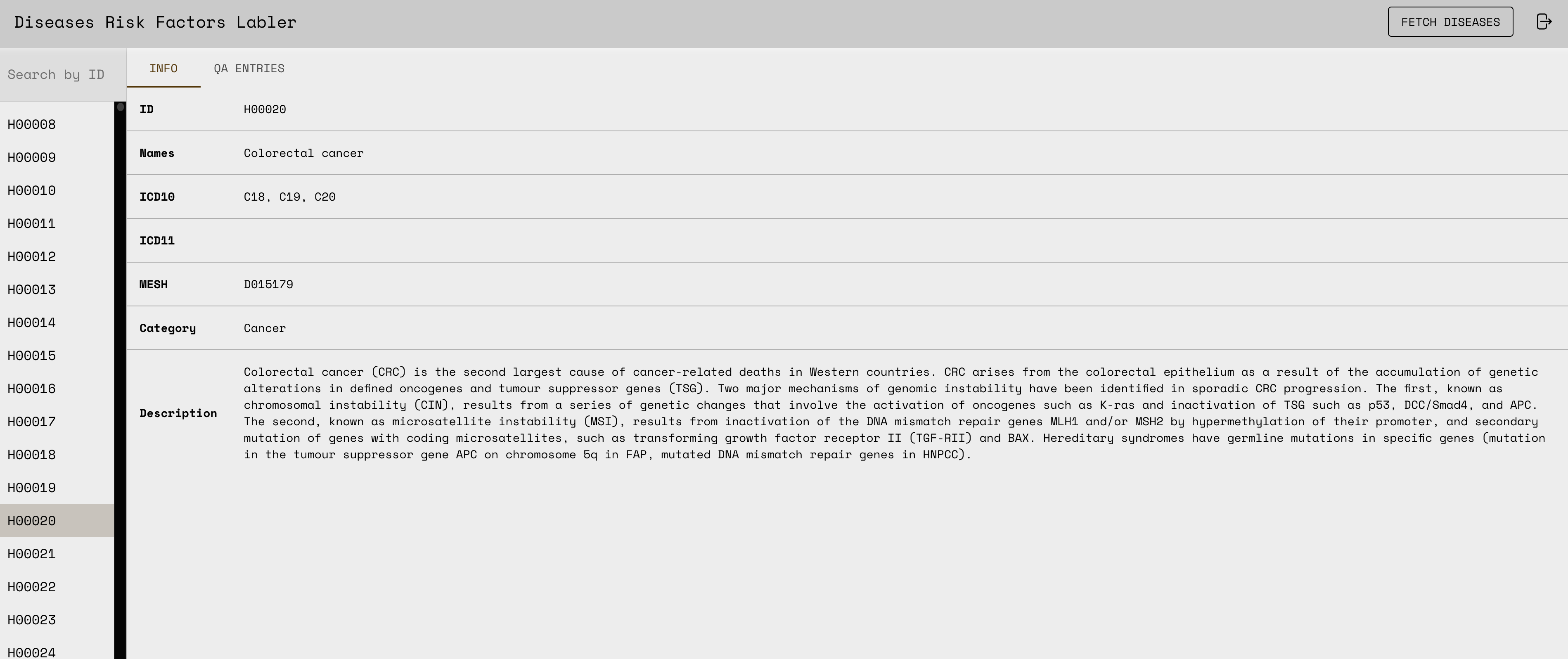}
\caption{Disease Risk Factor Annotation System: disease details as retrieved from KEGG and parsed.}
\label{fig:system1}
\end{figure*}

\begin{figure*}
\centering
\includegraphics[width=1\textwidth]{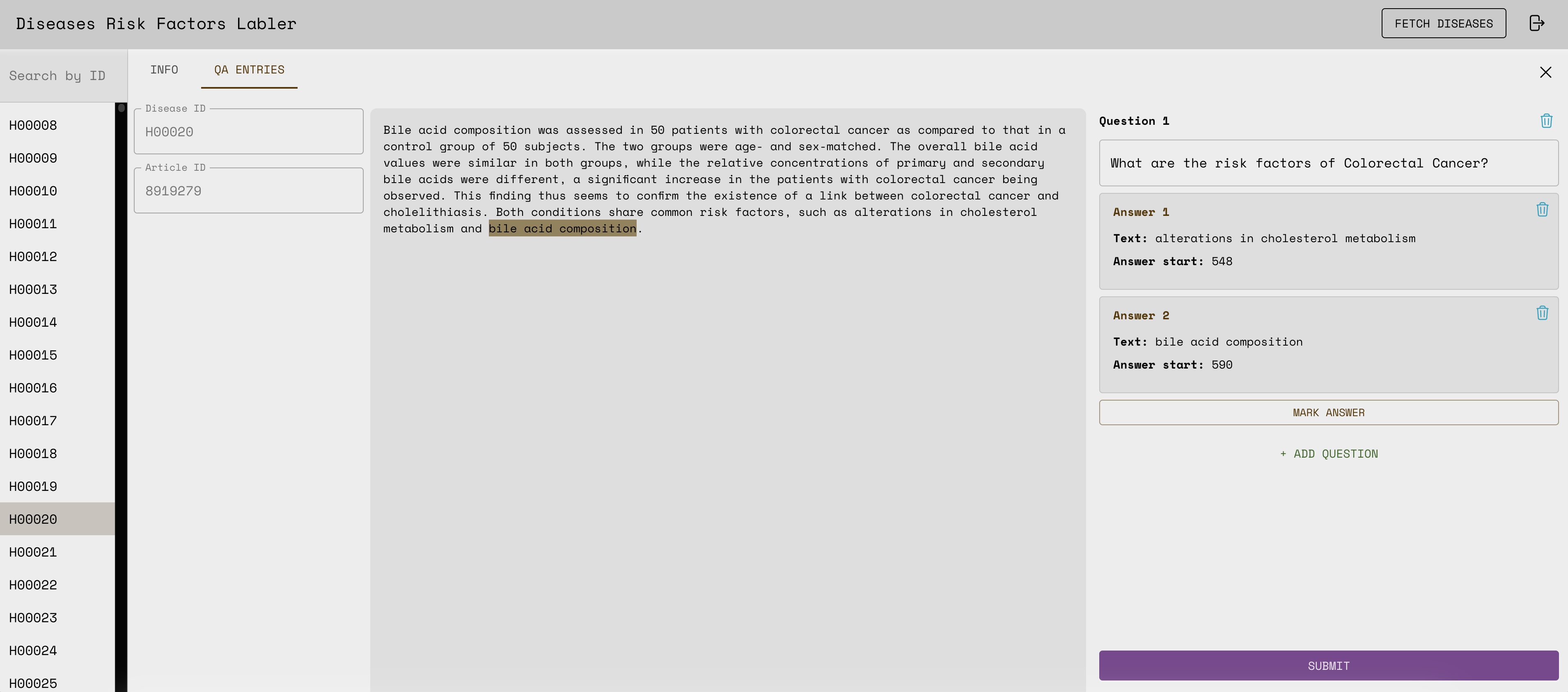}
\caption{Disease Risk Factor Annotation System: manual annotation of spans containing risk factors; multiple risk factors for the same disease can be identified in the same abstract.}
\label{fig:system2}
\end{figure*}

\paragraph{Python Algorithm}
Complementing the user interface is a Python-based algorithm that interacts with the GraphQL server. This component is responsible for processing medical articles, including sending requests to the server to fetch articles for annotation and submitting the results of automated risk factor identification processes. It plays a critical role in pre-processing and post-processing steps in the dataset creation pipeline.

\paragraph{Database}
At the core of the system lies a MongoDB database hosted on Azure Cosmos DB. This NoSQL database was chosen for its scalability, flexibility, and robust support for storing unstructured data, such as medical article texts and annotations. It stores all data related to diseases, articles, and user annotations, providing a persistent and reliable data storage solution for the system.

Figures~\ref{fig:system1}-\ref{fig:system2} illustrate two screenshots of the application developed for manual annotation of risk factors. The system code will also be made available per acceptance.

\section{Diseases with Annotated Risk Factors in the QA dataset (the training set)}
\label{sec:annotated-diseases}
Section~\ref{sec:rf-annotation} details the procedure of manual annotation of risk factors following the step of abstract retrieval. The annotated data comprises 1,712 QA items from 668 abstracts covering 15 diseases from multiple disease families, as detailed in Table~\ref{tbl:rf-families-ann}.

\begin{table}
\centering
\resizebox{\columnwidth}{!}{
\begin{tabular}{ll}  
family & disease \\ \hline
Autoimmune disease	& Celiac disease \\
Autoimmune disease & Rheumatoid arthritis \\
Autoimmune disease & Type 1 diabetes mellitus \\
Carcinomas	& Bladder cancer \\
Carcinomas (to the most part) &	Breast cancer \\
Carcinomas (to the most part) & Colorectal cancer \\
Chronic lung disease & Chronic obstructive pulmonary disease \\
Chronic lung disease & Asthma \\
Circulatory disorder & High blood pressure \\
Heart disease &	Myocardial infarction \\
Melanoma/Skin cancer & Melanoma \\
Metabolic disease & Metabolic syndrome \\
Metabolic disease & Type 2 diabetes mellitus \\
Neurodegenerative disorder	& Alzheimer disease \\
Neurologic disorder	& Migraine \\

\end{tabular}
}
\caption{Disease distribution by disease family in the manually annotated set of 1,712 risk factors used for BioBERT QA fine-tuning.}
\label{tbl:rf-families-ann}
\end{table}

\section{Diseases with Evaluated Risk Factors}

Table~\ref{tbl:rf-families-evl} reports the distribution of manually evaluated risk factors by disease family.

\begin{table}
\centering
\resizebox{\columnwidth}{!}{
\begin{tabular}{ll}  
family & disease \\ \hline
other hematological disorders & Multiple myeloma  \\
Carcinomas & Choriocarcinoma  \\
Carcinomas & Esophageal cancer  \\
Carcinomas & Gastric cancer  \\
Carcinomas & Malignant pleural mesothelioma  \\
Carcinomas & Non-small cell lung cancer  \\
Carcinomas & Penile cancer  \\
Carcinomas & Renal cell carcinoma  \\
Carcinomas & Small cell lung cancer  \\
Carcinomas & Vulvar cancer  \\
Infection & Cholera  \\
infection & Gonococcal infection  \\
infection & Pertussis  \\
Leukemias & Acute myeloid leukemia  \\
Leukemias & Adult T-cell leukemia  \\
Leukemias & B-cell acute lymphoblastic leukemia  \\
Leukemias & Chronic lymphocytic leukemia  \\
Leukemias & Chronic myeloid leukemia  \\
Leukemias & Hairy cell leukemia \\
Leukemias & Polycythemia vera  \\
Leukemias & T-cell acute lymphoblastic leukemia  \\
Lymphomas & Burkitt lymphoma  \\
Lymphomas & Lymphoplasmacytic lymphoma  \\
Metabolic disorders (GD) & Congenital adrenal hyperplasia  \\
Metabolic disorders (GD) & Gaucher disease  \\
Metabolic disorders (GD) & Hemochromatosis  \\
Mucus malefunction (GD) & Cystic fibrosis  \\
Cardiomyopathy & Dilated cardiomyopathy  \\
Sarcomas & Osteosarcoma  \\
\end{tabular}
}
\caption{Disease distribution by disease family in the manually evaluated set of 1,485 identified risk factors. "GD" denotes "genetic disorder".}
\label{tbl:rf-families-evl}
\end{table}

\end{document}